\documentclass[conference]{IEEEtran}
\IEEEoverridecommandlockouts

\usepackage{cite}
\usepackage{amsmath,amssymb,amsfonts}
\usepackage{algorithmic}
\usepackage{graphicx}
\usepackage{textcomp}
\usepackage{xcolor}
\usepackage{amsmath}
\usepackage{tabularx} 
\def\BibTeX{{\rm B\kern-.05em{\sc i\kern-.025em b}\kern-.08em
    T\kern-.1667em\lower.7ex\hbox{E}\kern-.125emX}}
\begin{document}

\title{Digitization of Analogue Gauges using Computer Vision\\}

\author{\IEEEauthorblockN{1\textsuperscript{st} Erico Kayumba Mutonkole}
\IEEEauthorblockA{\textit{dept. of Computer Science} \\
\textit{University of the Western Cape}\\
Cape Town, South Africa \\
3805086@myuwc.ac.za}
\and
\IEEEauthorblockN{2\textsuperscript{nd} Olasupo Ajayi}
\IEEEauthorblockA{\textit{dept. of Computer Science} \\
\textit{University of the Western Cape}\\
Cape Town, South Africa\\
ooajayi@uwc.ac.za}
\and
\IEEEauthorblockN{3\textsuperscript{rd} Omowunmi Isafiade}
\IEEEauthorblockA{\textit{dept. of Computer Science} \\
\textit{University of the Western Cape}\\
Cape Town, South Africa \\
oisafiade@uwc.ac.za}
}

\maketitle

\begin{abstract}
Analogue gauges remain prevalent in manufacturing industries for measuring physical quantities, yet manual monitoring by operators introduces human error and operational inefficiency. This research developed and evaluated an automated gauge reading system using computer vision and deep learning techniques. Three hybrid approaches were implemented and compared: OpenCV with Scikit-Image (traditional computer vision), OpenCV with ResNet18, and OpenCV with VGG19 (deep learning approaches). A comprehensive dataset of 1,848 pressure gauge images (0-10 BAR range) was curated through web scraping, synthetic generation, and augmentation techniques. The experimental results demonstrate that ResNet18 significantly outperforms the alternatives, achieving  an accuracy of 97. 3\%, compared to VGG19's 64.4\% and Scikit-Image's 63. 8\%. The study validates the effectiveness of residual learning architectures for industrial gauge digitization, providing a robust foundation for Industry 5.0 automation applications. 
\end{abstract}


\section{Introduction}
Analogue gauges are scientific instruments designed to measure physical quantities. These include temperature, mass and voltage to name a few. Despite being an older technology, analogue gauges are still widely used in many industries today to take measurement readings for various equipment \cite{b4}. However, it is easier for users to read digital numbers compared to reading from an analogue gauge thereby, reducing guesses of what value the meter pointer is pointing to\cite{b2}.

Manual gauge reading in industrial environments presents significant operational challenges and costs. Studies indicate that manual readings can have error rates of up to 5\%, often due to fatigue, distraction, or misinterpretations \cite{b2,b12}, while broader research on manual data entry shows typical error rates of about 1\%, with some studies reporting rates as high as 4\% for data entered without verification layers \cite{b13,b14}. In large industrial plants such as oil and gas refineries, manual reading tasks are very error-prone, time-consuming and often dangerous due to exposed positions of the gauges \cite{b5}. These errors translate into production inefficiencies, unnecessary process adjustments, and potential safety risks in critical industrial operations.

When we take cost into consideration, especially in industrial settings, rather than installing new digital meters, it might be cheaper to attach tiny cameras attached to cheap single board computers, such as Raspberry PI or ESP32, to capture images that will be processed by image recognition models, to obtain numeric readings \cite{b2}. Despite their practicality,  a comprehensive comparison between traditional computer vision approaches and modern deep learning architectures for gauge digitization on low resource devices remains limited in existing literature.

This work aims at developing a system that converts analogue readings to their digital equivalent using computer vision. The research also addresses the gap in literature by systematically comparing three distinct approaches: traditional computer vision (OpenCV with Scikit-Image), modern residual learning (OpenCV with ResNet18), and deep sequential learning (OpenCV with VGG19). The primary research objective is to determine whether architectural innovations in deep learning, particularly residual connections, provide significant performance advantages over traditional methods for analogue gauge classification tasks.
The OpenCV library will be used to preprocess images of analogue gauges, which will then be used to train the deep learning model. The trained model will accurately determine the numerical or digital equivalent value of the gauge’s reading.

\section{LITERATURE REVIEW}
A recent review on deep learning approaches for digital image processing \cite{b1} examined a wide range of techniques tailored for various application areas. The study provided a thorough investigation into several image processing tasks, including noise reduction, image enhancement, segmentation, feature detection, and classification. To evaluate the performance and effectiveness of these methods, several metrics were employed, such as mean squared error (MSE), peak signal-to-noise ratio (PSNR), structural similarity index (SSIM), mean structural similarity index (MSSIM), mean absolute error (MAE), and naturalness image quality evaluator (NIQE), among others. The review highlighted the strong noise reduction capabilities of methods like Self2Self neural networks, DnCNNs, and DFT-Net. In terms of classification, architectures ranging from Residual Networks to combinations like CNN-LSTM were recognized for their promising accuracy \cite{b1}.

In a study focused on water meter reading recognition using deep learning techniques \cite{b2}, three models namely Faster R-CNN, SSD, and YOLOv3 were evaluated to identify which offered superior performance. Among the models assessed, YOLOv3 achieved the highest recognition accuracy, recording a success rate of 90.61\%.

A proposed automated system combined deep learning, machine learning, and image processing to detect readings from analogue gauges \cite{b3}. The researchers utilized image processing techniques to generate supervised data from gauge images, addressing the challenges of working with unsupervised data and aiming to improve recognition accuracy. Three models—DenseNet 169, InceptionNet V3, and VGG19—were trained on 1,011 labeled images categorized into 9 classes with labels from 0 to 8. VGG19 showed a high training accuracy of 97\% but a lower testing accuracy of 75\%, indicating possible overfitting. In contrast, InceptionNet V3 showed consistent performance across both datasets, while DenseNet 169 delivered the highest accuracy and strongest generalization ability among the models  \cite{b3}.
 
In an effort to automate analogue gauge reading across various industries, including healthcare and automotive, researchers in \cite{b4} developed an algorithm called the Analogue Gauge Reader (AGR) using artificial intelligence methods. The algorithm was implemented with open-source tools such as OpenCV, NumPy, and PIL. Experimental results demonstrated a notable reduction in error rates following the application of image preprocessing techniques. Moreover, the algorithm proved adaptable to a range of analogue gauge types captured under varying lighting conditions. In testing, the approach achieved high accuracy and precision, ranging from 99\% to 100\%  \cite{b4}.

Analogue meters located in areas that were hard-to-access or hazardous locations proved to be challenging as measurements for various equipment in the manufacturing plant were needed. A solution was proposed in \cite{b5} to remotely monitor these analogue gauges in areas that are difficult to reach or hazardous using a combination of a microcontroller device and a camera. The process involved using the camera to capture gauge images, which were then processed on a local computer using a Convolutional Neural Network (CNN) with the CenterNet HourGlass104 model, segmentation of the pointer was done with the Hough Transform, while polar transformation was used to calculate the pointer angle. The value indicated by the gauge was determined based on the pointer's angle, and displayed on a web-accessible dashboard. The approach achieved an average relative error of 0.95\%, demonstrating good accuracy and robustness for industrial environments \cite{b5}.

A real-time mobile solution for reading circular single-pointer analogue gauges using a lightweight CNN-based system was presented in \cite{b6}. They used a synthetic dataset of 10,000 images with ground truth annotations to train the CNN model they implemented. The focus of the research was to implement an efficient method for processing both high- and low-resolution gauge images on mobile phones, and to the release of a diverse real-world gauge dataset. They achieved pointer angle errors below 1° and while running at up to 25fps on mobile devices \cite{b6}.

\section{RESEARCH FOCUS}
This work addresses the fundamental challenge of automated interpretation of analogue gauges using computer vision. As a proof of concept and to demonstrate the viability of this idea, only pressure gauges of up to 10 BAR were considered. The research investigates three distinct hybrid approaches: OpenCV with Scikit-Image (traditional computer vision), OpenCV with ResNet18 (modern residual learning), and OpenCV with VGG19 (deep sequential architecture). This comparative framework enables systematic evaluation of traditional versus deep learning methodologies while examining the impact of architectural innovations (residual connections) on classification performance. The choice of these approaches is grounded in their theoretical significance. Scikit-Image represents traditional computer vision techniques, VGG19 illustrates deep sequential architectures that are susceptible to vanishing gradient issues \cite{b18}, and ResNet18 demonstrates modern advancements designed to overcome gradient flow challenges. Their performance is evaluated using accuracy, precision, recall, F1-score and mean average precision (mAP), providing a comprehensive assessment across multiple dimensions.

\section{REQUIREMENT ANALYSIS}
Computer vision has vast industrial applications and has steadily been growing in popularity in recent times. This work’s scope is limited to image recognition, image processing and deep learning models and their applications in digitizing analogue gauges. 

To be considered practical and viable, the proposed solution must meet the following functional requirements: 
Images of analogue gauge must be acquired from camera sensors.
\begin{enumerate}
  \item Computer vision models should be used to analyse the captured image to obtain the equivalent digital reading.
  \item Digital value inference must have a precision targeting ±0.1 BAR accuracy
  \item Real-time processing capability for industrial applications, with approximately 2 seconds processing time per image.
  \item The developed system should be able to work on low-spec hardware such as microcontrollers or single board computers
\end{enumerate}

\section{METHODOLOGY AND EXPERIMENTAL DESIGN}
Analogue gauge digitization is formulated as a multi-class classification problem where each discrete gauge reading (0-10 BAR) represents a distinct class. This approach, while discretizing continuous readings, provides robust performance against visual noise and simplifies the mapping from visual input to numerical output compared to regression-based needle angle estimation methods\cite{b20}. The following sub-sections discuss our research methodology.

\subsection{Data Collection and Preparation}
The data for this experiment were prepared as follows:
\begin{enumerate}
    \item A custom dataset was curated by web scraping pressure gauge images from Google Images. This was necessary because, for the task of analogue gauge reading, there is a significant lack of large, publicly available datasets, a limitation also acknowledged by\cite{b6}. Three (3) types of pressure gauges were selected as shown in Figure 1. These gauges had readings from 0 to 10 BAR (11 classes).
    \item The 3 images were then synthesized to generate 33 image classes using Adobe Photoshop. 11 images per class * 3 = 33. A sample of the synthesized image is shown in Figure 2.
    \item Finally, we augmented the resulting 33 images using various techniques and ended up with 1,848 augmented pressure gauge image variants.
\end{enumerate}
 
\begin{figure}[htbp]
\centerline{\includegraphics[width=\linewidth]{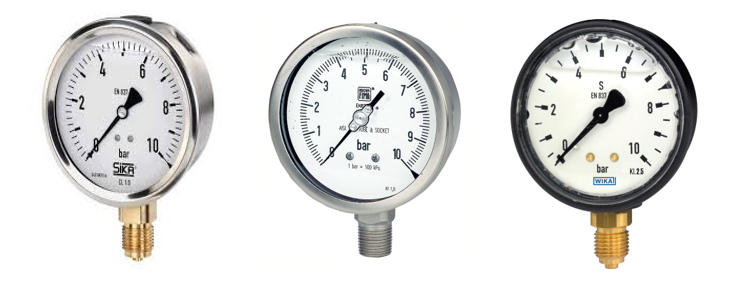}}
\caption{Three original gauges sourced from Google Images.}
\label{fig}
\end{figure}

\begin{figure}[htbp]
\centerline{\includegraphics[width=\linewidth]{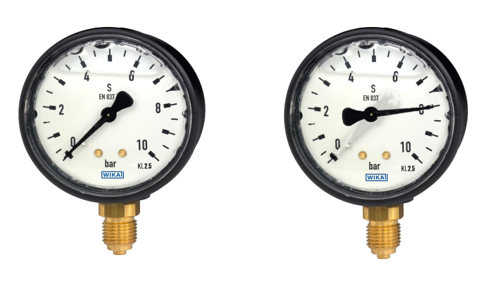}}
\caption{Original (Left) and synthetic (right) image.}
\label{fig}
\end{figure}

\subsubsection{Data augmentation pipeline}
The augmentation process was designed to improve model robustness to real-world variations in gauge imaging. Five augmentation processes were carried out.
\begin{enumerate}
    \item The images were randomly rotated by ±15° to simulate different camera angles.
    \item The images were scaled between 90\% and 110\% to account for changes in camera distance.
    \item Positional robustness was introduced through translation of up to 10\% of the image’s width and height along both axes. 
    \item Perspective transformations were applied to mimic changes in viewing angles.
    \item Salt-and-pepper noise injection was used to simulate sensor noise and environmental artifacts.
\end{enumerate}

\begin{figure}[htbp]
\centerline{\includegraphics[width=\linewidth]{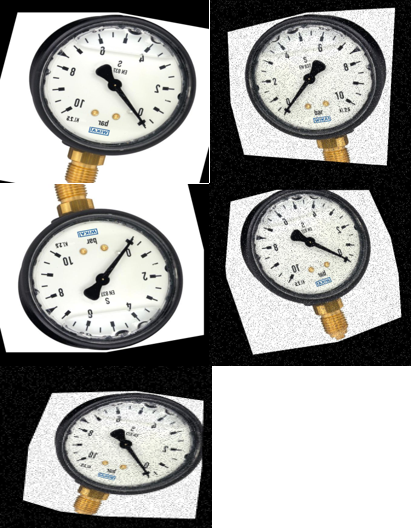}}
\caption{Samples of augmented images.}
\label{fig}
\end{figure}

\subsubsection{Dataset Partitioning Pipeline}
The dataset was then split into an 80:20 ratio. 80\% for training (80\%, n=1.474) and 20\% for Validation (20\%, n=374).

\subsection{Image Processing}
The following steps were taken to process the images in the dataset:
\begin{enumerate}
    \item Conversion to Grayscale
    
The images were converted to grayscale. Removing colour information simplifies the images and enhances the intensity. This step has been shown to help in reducing computational complexity and is essential for edge detection and further analysis, as reported in [7]. A comparison of the original vs gray scale version is shown in Figure 4

\begin{figure}[htbp]
\centerline{\includegraphics[width=\linewidth]{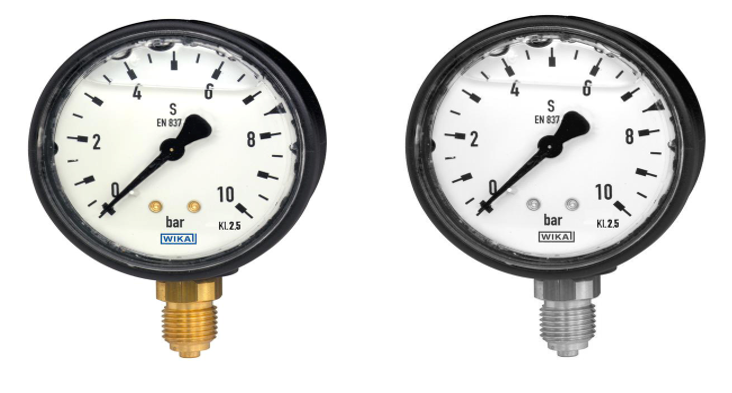}}
\caption{Original (left) image and Grayscale (right) image.}
\label{fig}
\end{figure}

\item Extraction of Region of Interest (RoI)

We isolated the gauge area from the background to focus on the circular dial of the gauge and filter out irrelevant parts of the image.
\item Gauge Labelling 

Traditionally, when using computer vision for  analogue gauge digitization, the model estimates the needle's angle to determine the reading as documented in\cite{b21}. In contrast, this study used a classification-based approach, treating readings as discrete labeled values. This ensured that our computer vision models would be more robust to visual noise and simplify the mapping from image to value. 
To achieve this we framed this work as an image classification problem and utilized deep learning models (ResNet18 and VGG19). Being supervised learning models, we had to first manually label the images, assigning each gauge image a class label that represents its reading. For instance, if the gauge’s needle points to '6', we  put it in class '6'. 
\end{enumerate}

\subsection{Models}
As stated earlier, we explored and compared 3 hybrid approaches in this work, namely, OpenCV with Scikit-Image, OpenCV with ResNet18, and OpenCV with a VGG19. PyTorch framework was used to train and evaluate the deep Learning models.

\subsubsection{OpenCV with Scikit-Image}
This approach used OpenCV to load and preprocess images and then applied scikit-image for enhanced filtering and region analysis [8]. Scikit-Image is a Python library built on top of NumPy, SciPy, and matplotlib. We tagged this the “Traditional computer vision technique” as it employed a four-stage approach for needle angle detection and classification. The stages are summarized as follows:
\begin{itemize}
    \item Canny edge detection with low and high thresholds of 50 and 150 respectively, was used to identify the needle boundaries.
    \item Probabilistic Hough Line Transform was then used to extract the needle line from the detected edges.
    \item Geometric computation was used to calculate the needle’s angle relative to the gauge center. 
    \item The obtained angles were mapped  to discrete classes using predefined angular intervals of 32.7 (360 divided by 11) degrees per class, enabling classification across 11 classes spanning the full 360-degree range. This systematic approach helped in uniform distribution across the entire angular spectrum, with each class representing an equal portion of the gauge’s measurement range. 
\end{itemize}

\subsubsection{OpenCV with ResNet18}
This hybrid approach leveraged OpenCV and ResNet18 (implemented in PyTorch). OpenCV was used for image loading and preprocessing, specifically for image resizing and conversion to grayscale, while ResNet18 was used for classification. Our implementation process can be summarized as follows: 
\begin{itemize}
    \item The images were converted to  to 224x224 pixels and converted to 3-channel grayscale format to meet the input requirements of pre-trained ResNet18.
    \item We utilized apre-trained ResNet18 backbone for feature extraction, but still trained it on our dataset. By utilizing a pre-trained model we were able to minimize the training time and benefitted from transfer learning \cite{b18}. We froze the ResNet model during the first initial 10 epochs before being fine-tuned in subsequent training phases. In total we trained the model using 100 epochs. 
    \item The classification component of our model consisted of a modified final fully connected layer that mapped the 512-dimensional feature representations to 11 output classes. We applied a dropout rate of 0.3 before the final layer to mitigate overfitting and improve generalization performance. 
\end{itemize}

\subsubsection{OpenCV with VGG19}
Like in the previous model, we also applied OpenCV to preprocess the images and convert them to 224×224×3 RGB images. The converted images were then passed as input to a VGG19 model pre-trained on ImageNet for feature extraction.
\begin{itemize}
    \item The original classifier structure of 25088→4096→4096→1000 was modified to 25088→1024→512→11 to accommodate the 11-class classification task.
    \item We applied ReLU activation functions and dropout layers throughout the network. Finally, to prevent overfitting we applied a dropout rate of 0.5 across each of the fully connected layers in the modified classifier.
\end{itemize}

\subsubsection{Performance Metrics}
To evaluate the performance of the model, the following metrics were used:
\begin{itemize}
    \item Accuracy -  Measures how close the digitized value was to the true pressure value (ground truth). 
    
    \begin{equation}
    \text{Accuracy} = \frac{TP + TN}{TP + TN + FP + FN}
    \end{equation}\cite{b15}
    
    \item Precision - Measures the consistency of repeated digital equivalent values of the same pressure value of the analogue gauge.

    \begin{equation}
    \text{Precision} = \frac{TP}{TP + FP}
    \end{equation}\cite{b15}

    \item Recall - This measured the model’s ability to detect and return the correct digital equivalent value with respect to the pressure value across the entire gauge range.

    \begin{equation}
    \text{Recall} = \frac{TP}{TP + FN}
    \end{equation}\cite{b15}

    \item F1-Score -  This computed the harmonic mean of precision and recall for the models to ensure that all readings are captured and there are no missed readings.

    \begin{equation}
    F_1 = 2 \cdot \frac{\text{Precision} \cdot \text{Recall}}{\text{Precision} + \text{Recall}}
    \end{equation}\cite{b15}

    \item Cross-Entropy Loss - This  measured the average difference between predicted probability distribution and true distinct values (labels) readings of the analogue gauges.

    \begin{equation}
    \mathcal{L}_{\text{CE}} = - \frac{1}{N} \sum_{i=1}^{N} \sum_{c=1}^{C} y_{i,c} \log(\hat{y}_{i,c})
    \end{equation}\cite{b16}

    \item Mean Average Precision (mAP) - This was used as the primary evaluation metric to assess the performance of the hybrid approaches across all gauge values. It is defined as:

    \begin{equation}
    AP_i = \int_0^1 P_i(R) \, dR
    \end{equation}\cite{b19}
    
    \begin{equation}
    \text{mAP} = \frac{1}{N} \sum_{i=1}^{N} AP_i
    \end{equation}\cite{b19}
    
\end{itemize}

\subsection{Visualization and Analysis Tools}
The evaluation methodology encompasses comprehensive performance assessment through multiple visualization and analytical techniques, including confusion matrices to examine class-wise performance and identify misclassification patterns, ROC curves with corresponding AUC values to evaluate binary classification performance for each individual class, training dynamics plots that display loss curves and accuracy progression throughout the training process.

\section{RESULTS AND ANALYSIS}
The performance of the three proposed approaches implemented in this work are presented in this section.
\subsection{Comparative Performance Analysis}
We start off by comparing ResNet18 and VGG19 as they are both deep learning models. This comparison is important because it highlights the relative strengths and weaknesses of modern convolutional neural network architectures when applied to the task of analogue gauge digitization. ResNet18, with its residual connections, is designed to address the vanishing gradient problem and typically achieves higher accuracy with fewer parameters, while VGG19, although deeper, is computationally more intensive and less efficient in feature reuse. Evaluating these models side by side provides insight into the suitability of deep learning–based feature extraction approaches for this domain.
In contrast, the subsequent comparison with Scikit-Image will be based on overall performance rather than per-class metrics. This is because Scikit-Image employs traditional computer vision techniques that rely on handcrafted features rather than learned representations. Such methods do not produce class-wise probabilities in the same way as deep learning classifiers; instead, their effectiveness is better reflected in aggregate accuracy or error rates across the dataset. As a result, reporting overall performance provides a fairer and more meaningful basis for evaluating the relative strengths of classical image processing versus deep learning–based approaches in this application.

\begin{table}[htbp]
\centering
\setlength{\tabcolsep}{2pt}
\caption{Comparison of ResNet18 and VGG19}
\small
\begin{tabularx}{\columnwidth}{|c|cc|cc|cc|}
\cline{1-7}
\textbf{Class} &
\multicolumn{2}{|c|}{\textbf{Precision}} &
\multicolumn{2}{|c|}{\textbf{Recall}} &
\multicolumn{2}{|c|}{\textbf{F1-Score}} \\
\cline{2-7}
 & \textbf{ResNet18} & \textbf{VGG19}
 & \textbf{ResNet18} & \textbf{VGG19}
 & \textbf{ResNet18} & \textbf{VGG19} \\
\cline{1-7}
0  & 1.00 & 0.57 & 0.97 & 0.97 & 0.97 & 0.99 \\
1  & 0.72 & 1.00 & 0.94 & 0.92 & 0.97 & 0.35 \\
2  & 0.96 & 0.51 & 0.77 & 0.40 & 0.97 & 0.97 \\
3  & 0.86 & 0.57 & 1.00 & 0.69 & 0.94 & 0.65 \\
4  & 0.97 & 0.68 & 0.94 & 0.68 & 0.96 & 0.68 \\
5  & 0.94 & 0.79 & 1.00 & 0.68 & 0.97 & 0.73 \\
6  & 1.00 & 0.67 & 0.88 & 0.82 & 0.94 & 0.74 \\
7  & 0.97 & 1.00 & 0.97 & 0.35 & 0.97 & 0.52 \\
8  & 0.96 & 0.74 & 0.77 & 0.59 & 0.85 & 0.66 \\
9  & 0.92 & 0.43 & 0.97 & 0.18 & 0.94 & 0.25 \\
10 & 0.94 & 0.94 & 0.97 & 0.85 & 0.96 & 0.89 \\
\cline{1-7}
\end{tabularx}
\label{tab:metrics}
\end{table}

Table 1 shows that ResNet18 consistently outperformed the VGG19 model in almost all classes in terms of precision, recall and F1-score metrics. The ResNet18 model achieved a near-perfect performance on classes 0, 3, 5, and 7, with  F1-score $\geq$ 0.97, demonstrating its robustness in distinguishing fine-grained variations in pointer positions. In contrast,VGG19 was inconsistent, performing strongly in some classes (like class 10 with F1-score of 0.89), but poorly in others (like class 9 with F1-score of 0.25). These results show a significant performance disparity between ResNet18 and VGG19, indicating that ResNet18 was able to generalise and classify readings with a high precision, recall and F1-score. 

Figure 5 and 6 further indicate the performance disparity between ResNet18 and VGG19 models. Figure 6 shows that VGG19 exhibited widespread misclassification patterns with significant off-diagonal confusion, especially between classes 2 - 4 and scattered predictions for classes 0, 3, 4, 9, and 10.  This dispersion indicates that VGG19 struggles to learn class-specific representations, leading to inconsistent recall across multiple categories. Conversely, ResNet18, depicted  in Figure 5, demonstrated a near perfect classification with a strong diagonal pattern indicating minimal confusion between classes with the majority of predictions concentrated along the correct class axis. Misclassifications were minimal and largely limited to a few neighboring classes, such as 5 being confused with 6 or 8 being confused with 9. Such confusions are justifiable, as even humans might also confuse 5 and 6 or 8 and 9, especially when cursive / calligraphy fonts are used. In summary, ResNet18 is able to generalize effectively while maintaining high precision and recall across most categories. 

\begin{figure}[htbp]
\centerline{\includegraphics[width=\linewidth]{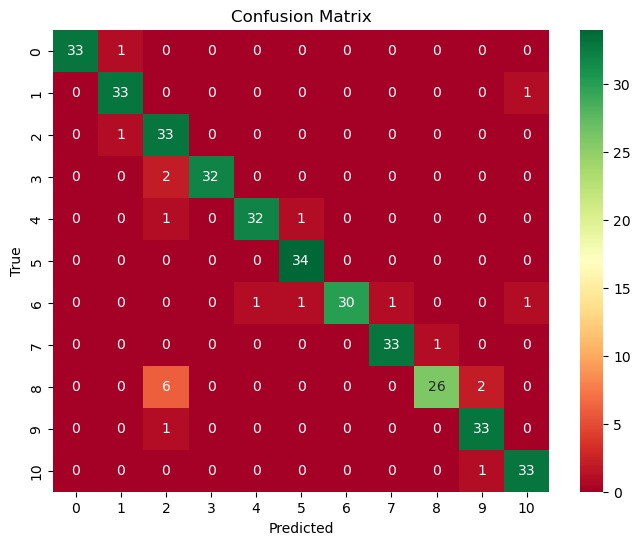}}
\caption{Confusion matrix for ResNet18.}
\label{fig}
\end{figure}

\begin{figure}[htbp]
\centerline{\includegraphics[width=\linewidth]{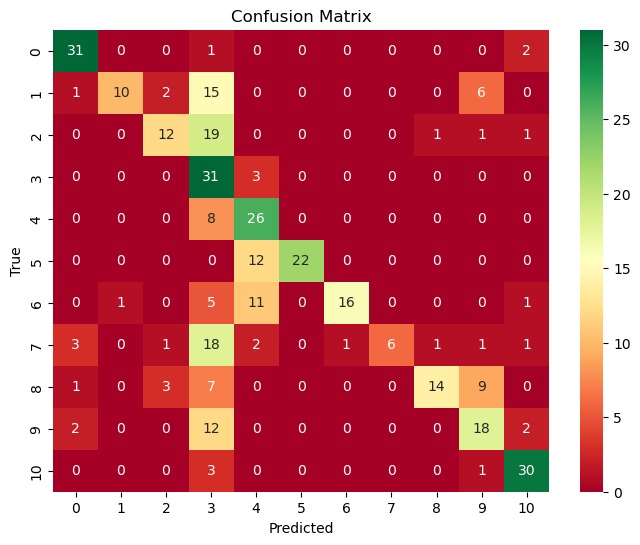}}
\caption{Confusion matrix for VGG19.}
\label{fig}
\end{figure}

\begin{figure}[htbp]
\centerline{\includegraphics[width=\linewidth]{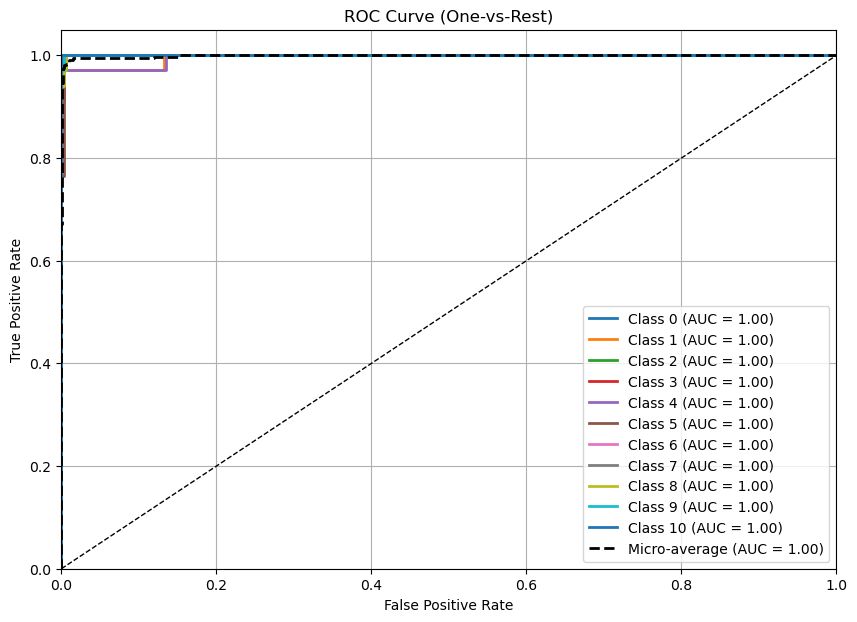}}
\caption{ROC Curve for ResNet18.}
\label{fig}
\end{figure}

\begin{figure}[htbp]
\centerline{\includegraphics[width=\linewidth]{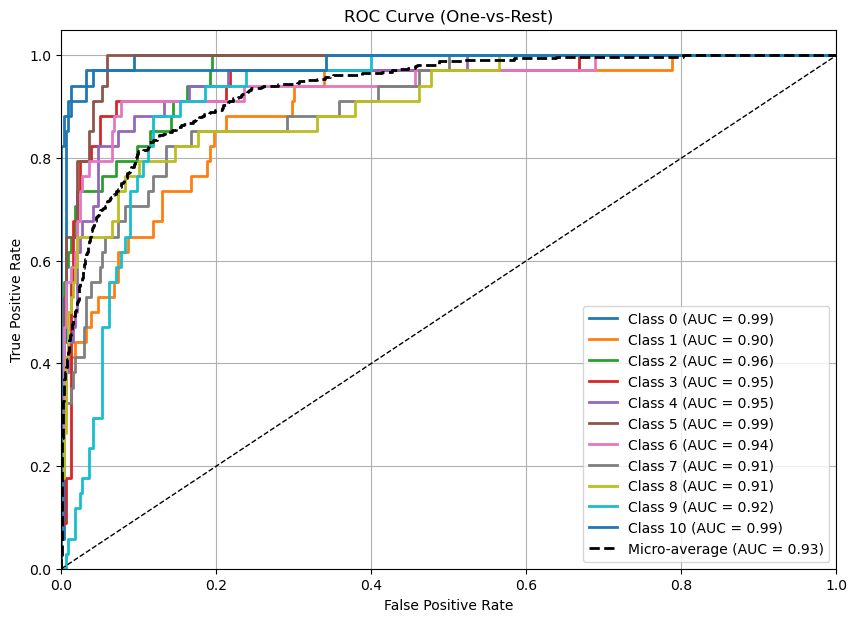}}
\caption{ROC Curve for ResNet18.}
\label{fig}
\end{figure}

The ROC curves, shown in Figures 7 and 8, further reinforce the differences observed between VGG19 and ResNet18. For VGG19, while the overall area under the curve (AUC) values appear close to 1.0 for all classes, the curve shapes indicate a degree of instability in classification confidence, which aligns with the misclassifications evident in the confusion matrix (Figure 6). In contrast, ResNet18 consistently achieved near-perfect AUC values of 1.0 across all classes, with the curves tightly hugging the top-left corner of the plot. This indicates near-perfect separation between true positives and false positives, reflecting both high sensitivity and specificity. 

\subsection{Error Analysis}
\begin{figure}[htbp]
\centerline{\includegraphics[width=\linewidth]{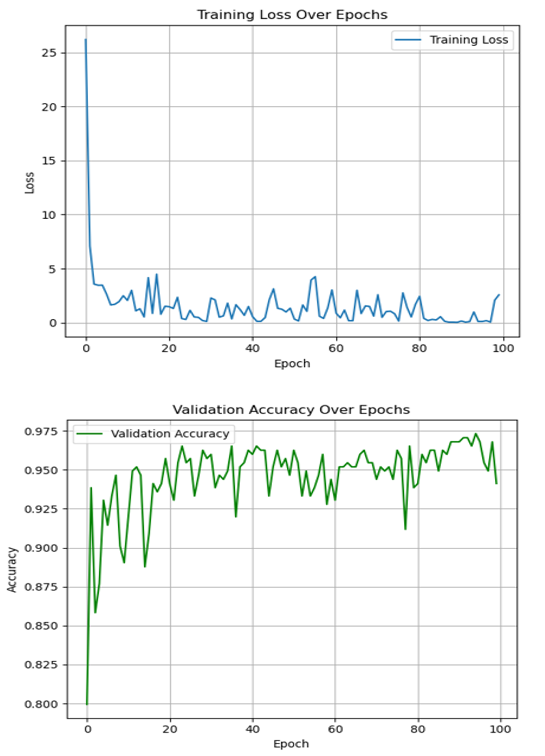}}
\caption{Training loss and Validation Accuracy over epochs for ResNet18.}
\label{fig}
\end{figure}

\begin{figure}[htbp]
\centerline{\includegraphics[width=\linewidth]{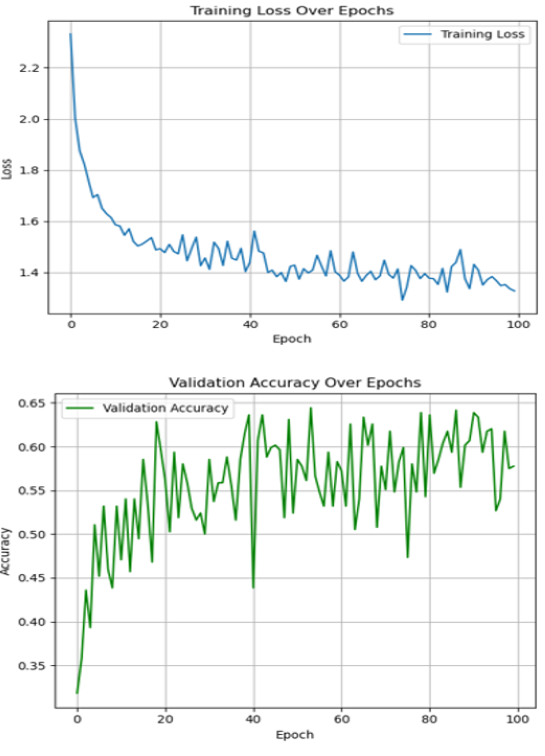}}
\caption{Training loss and Validation Accuracy over epochs for VGG19.}
\label{fig}
\end{figure}

Figures 10 and 11 display the training dynamics between ResNet18 and VGG19, revealing fundamental differences in their learning capabilities and optimization characteristics. ResNet18 converged with its training loss rapidly decreasing from approximately 25 to near-zero within the first 20 epochs. This was followed by a stable low-loss training throughout the remaining epochs. Correspondingly, its validation accuracy quickly reaches and maintains 95-97\% with relatively minor fluctuations, indicating robust generalization and effective learning. VGG19, on the other hand, plateaued around 1.4 - 1.6 after initial descent, suggesting convergence to a suboptimal local minimum as its validation accuracy stagnates at only 55-65\% with high volatility, indicating poor generalization and potential overfitting despite the higher training loss.
While the comparison between ResNet18 and VGG19 highlighted the superiority of ResNet18 at the class level, it is equally important to contrast deep learning approaches with traditional image processing techniques such as Scikit-Image. Unlike neural networks, which learn representations from data, Scikit-Image relies on handcrafted feature extraction methods and classical algorithms to interpret gauge readings \cite{b17}. As a result, its performance is best evaluated in terms of overall accuracy and error rates rather than per-class breakdowns, since it lacks the adaptive capacity to handle nuanced variations across individual classes (which would have been shown with a confusion matrix). This broader evaluation provides valuable insights into whether the added complexity and computational cost of deep learning models are justified by their gains in precision, recall, and robustness when compared to a lightweight traditional approach. 
The ResNet18 approach demonstrated exceptional performance across all metrics, having achieved near-perfect accuracy (97.3\%) and mAP (99.1\%). This indicated robust feature learning and generalization capabilities of the model through transfer learning from the pre-trained ImageNet model. Despite having shown perfect precision, the Scikit-Image approach demonstrated significantly lower recall (63.8\%), suggesting that it was conservative in making predictions when it classified the classes, it was always correct, but it missed many actual cases, such that the method only successfully identified and classified 63.8\% of the actual needle positions presented in the test dataset. This conservative behavior is typical of traditional computer vision methods with strict thresholding [18].  Despite being a deep learning approach, VGG19 performed poorly compared to ResNet18. It achieved a slightly higher accuracy (64.4\%) compared to the Scikit-Image approach. However, VGG19 performed poorly in precision, recall, F1-score and mAP compared to the other two approaches (OpenCV with ResNet18 and OpenCV with Scikit-Image).

Table 2 presents a summarised comparison between the 3 approaches. The mAP was used to obtain the results on the table. 

\begin{table}[htbp]
\centering
\setlength{\tabcolsep}{3pt}
\caption{Comparison of ResNet18 vs VGG19 vs Scikit-Image}
\begin{tabularx}{\columnwidth}{|X|c|c|c|c|c|}
\hline
\textbf{Approach} &
\textbf{Accuracy} &
\textbf{Precision} &
\textbf{Recall} &
\textbf{F1-score} &
\textbf{mAP} \\
\hline
OpenCV + ResNet18     & 97.3\% & 94.7\% & 94.1\% & 94.1\% & 99.1\% \\
OpenCV + VGG19        & 64.4\% & 71.1\% & 64.4\% & 63.0\% & 75.6\% \\
OpenCV + Scikit-Image & 63.8\% & 100\%  & 63.8\% & 77.9\% & 100\% \\
\hline
\end{tabularx}
\label{tab:approaches}
\end{table}

\subsection{Computational Performance}
We also compared the computational resource and time required by the three models, as summarized in Table 3.

\begin{table}[htbp]
\centering
\setlength{\tabcolsep}{3pt}
\caption{Resource Usage and Efficiency Comparison}
\begin{tabularx}{\columnwidth}{|X|c|c|c|}
\hline
\textbf{Metric} &
\textbf{ResNet18} &
\textbf{VGG19} &
\textbf{Scikit-Image} \\
\hline
Training Time (hours) & 5 & 8 & N/A \\
Inference Time (ms)   & $23.4 \pm 2.1$ & $41.2 \pm 3.8$ & $127.5 \pm 15.2$ \\
Memory Usage (MB)     & 145 & 289 & 23 \\
Model Size (MB)       & 42.7 & 532.6 & N/A \\
\hline
\end{tabularx}
\label{tab:efficiency}
\end{table}

The table shows that ResNet18 took about 46\% less time to train than VGG19. Training times are not available for Scikit-Image because it is not a neural network model, hence does not require training. 

When exposed to new images, ResNet18 was almost twice as fast as VGG19 in performing inference (i.e., classifying the image). Scikit-Image, being based on numerical computations, was extremely slow in carrying out this task. It however made up for it with its very small memory footprint, requiring 6x and 12x less memory space than ResNet18 and VGG19 respectively. 

\section{DESIGN AND PROTOTYPING}
In this section we discuss potential design and implementation of a system that can leverage our proposed models. 

\begin{figure}[htbp]
\centerline{\includegraphics[width=\linewidth]{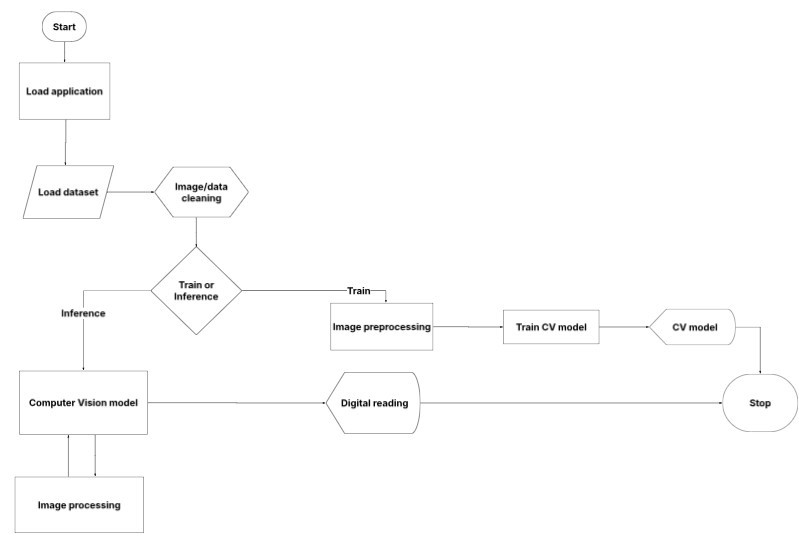}}
\caption{FlowChart of the system.}
\label{fig}
\end{figure}

\subsection{System Flow}
The flowchart in Figure 11 illustrates the process of digitizing analogue gauges using a computer vision model. It begins with loading the application and dataset, followed by image cleaning. Depending on the mode (training or inference), the workflow branches out either into a Training mode - where the cleaned dataset undergoes image preprocessing before training the computer vision (CV) model; Inference mode where preprocessed images are fed directly into the trained CV model, which performs inference and displays the obtained digital readings.The process loops back if needed or stops upon completion.

Note that, in this context, the term “computer vision or CV” refers to any of our proposed models (OpenCV + ResNet18, OpenCV + VGG19, or OpenCV + Scikit-Image). 

The workflow supports iterative improvement, such that, if inference results are unsatisfactory, the system can loop back to training with additional data. This pipeline accommodates both traditional CV techniques and deep learning, balancing accuracy and computational efficiency.

Implementation wise, the proposed design can be deployed on a mobile device or single board computer with an attached camera to capture images of the analogue gauges. 
\subsection{Interface Design}
A medium fidelity prototype was designed using Penport, an internet based prototyping tool [10]. Snapshots of the prototype as shown in Figure 12. 

Users need to be authenticated to use the system, as shown in Figure 12 (top left image). On first use, users are encouraged to select their desired approach to get inference under model. This can be done via the Settings page shown in the top right image of Figure 12.  

When an image of the analogue gauge has been successfully captured in the system Figure 12 (lower left image),  the user can then press the analyze button which will run the selected approach and return the digital equivalent value, as shown in the lower right image in Figure 12. 

\begin{figure}[htbp]
\centerline{\includegraphics[width=0.75\linewidth]{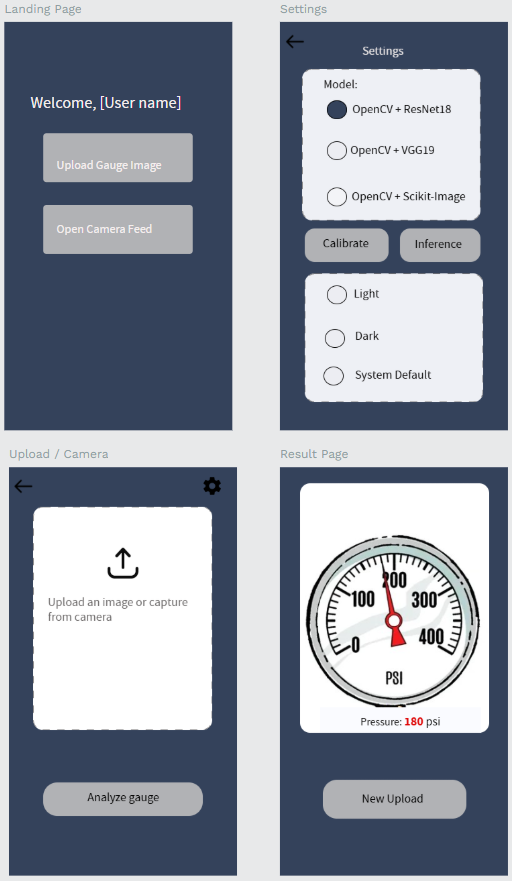}}
\caption{Medium-fidelity Prototype design .}
\label{fig}
\end{figure}

\section{CONCLUSION}
This study demonstrated the feasibility of digitizing analogue pressure gauges using computer vision. Three hybrid approaches were considered in this work. A traditional CV approach, "‘OpenCV with Scikit-image’, and two contemporary models based on deep learning, "‘OpenCV with ResNet18’ and ‘OpenCV with VGG19’. For all three approaches, OpenCV was used for image preparation, while the core model was used for inference. We used the three approaches to digitize 10 BAR analogue pressure gauges and compared their performance. The results obtained show that ResNet18 clearly outperformed both VGG19 and the traditional Scikit-Image-based model. The OpenCV+ResNet18 model achieved an average precision of 97.3\%and an average F1-score of 94.1\% across all classes, compared to OpenCV+VGG19’s 64.4\%
63\% and OpenCV+Scikit-Image’s 63.8\% and 77.9\%. 
 Confusion matrix and ROC analyses also buttressed ResNet18’s robustness, with near-perfect class separation. Though Scikit-Image offered high precision, its limited recall and lower overall performance highlight the insufficiency of classical image processing for robust gauge digitization. These findings establish deep learning, and specifically ResNet18, as a practical and effective solution for industrial monitoring, enabling accurate, automated gauge readings, minimizing human error, and supporting Industry 5.0 objectives through predictive maintenance and data-driven decision-making.

 While this study focused on 10-BAR pressure gauges, the methodology shows strong potential for broader applicability. With targeted fine-tuning, the pretrained ResNet18 and VGG19 models could be adapted to other analogue instruments such as temperature, voltage, or flow meters, opening pathways for unified, automated instrumentation monitoring across multiple domains. Overall, the results establish a solid foundation for future research on transferable gauge-reading systems and mark an important step toward intelligent, data-driven industrial environments.


\begin{thebibliography}{00}

\bibitem{b1} R. Archana and P. E. Jeevaraj, ``Deep learning models for digital image processing: a review,'' \textit{Artificial Intelligence Review}, vol. 57, no. 1, p. 11, 2024.

\bibitem{b2} Y. Liang, Y. Liao, S. Li, W. Wu, T. Qiu, and W. Zhang, ``Research on water meter reading recognition based on deep learning,'' \textit{Scientific Reports}, vol. 12, no. 1, p. 12861, 2022.

\bibitem{b3} H. Ninama, J. Raikwal, A. Ravuri, D. Sukheja, S. K. Bhoi, N. Z. Jhanjhi, \textit{et al.}, ``Computer vision and deep transfer learning for automatic gauge reading detection,'' \textit{Scientific Reports}, vol. 14, no. 1, p. 23019, 2024.

\bibitem{b4} S. Chavan, X. Yu, and J. Saniie, ``High precision analog gauge reader using optical flow and computer vision,'' in \textit{Proc. IEEE Int. Conf. Electro Information Technology (eIT)}, pp. 171--175, May 2022.

\bibitem{b5} J. Peixoto, J. Sousa, R. Carvalho, G. Santos, J. Mendes, R. Cardoso, and A. Reis, ``Development of an analog gauge reading solution based on computer vision and deep learning for an IoT application,'' \textit{Telecom}, vol. 3, no. 4, pp. 564--580, Oct. 2022.

\bibitem{b6} B. Howells, J. Charles, and R. Cipolla, ``Real-time analogue gauge transcription on mobile phone,'' in \textit{Proc. IEEE/CVF Conf. Comput. Vis. Pattern Recognit.}, pp. 2369--2377, 2021.

\bibitem{b7} C. Kanan and G. W. Cottrell, ``Color-to-grayscale: does the method matter in image recognition?,'' \textit{PLoS One}, vol. 7, no. 1, p. e29740, 2012.

\bibitem{b8} S. Van der Walt, J. L. Schönberger, J. Nunez-Iglesias, F. Boulogne, J. D. Warner, N. Yager, \textit{et al.}, ``scikit-image: image processing in Python,'' \textit{PeerJ}, vol. 2, p. e453, 2014.

\bibitem{b9} A. Krizhevsky, I. Sutskever, and G. E. Hinton, ``Imagenet classification with deep convolutional neural networks,'' in \textit{Advances in Neural Information Processing Systems}, vol. 25, 2012.

\bibitem{b10} Penpot, ``DAGCV, Penpot Design Workspace,'' 2024. [Online]. Available: https://design.penpot.app/\#/workspace?team-id=921945d0-8932-8162-8006-2492c33772c7\&file-id=95aea877-d515-80ac-8006-24932424a647\&page-id=95aea877-d515-80ac-8006-1f7b16b956f8

\bibitem{b11} X. Yang, ``Machine Learning in Practice: Using Artificial Intelligence to Read Analog Gauges,'' Object Computing, St. Louis, MO, USA, Jun. 2019. [Online]. Available: https://objectcomputing.com/resources/publications/sett/june-2019-using-machine-learning-to-read-analog-gauges

\bibitem{b12} Third Eye Data, ``The Hidden Cost of Manual Gauge Readings: Explaining the AI Solutions for Manufacturing Industry,'' May 6, 2024. [Online]. Available: https://thirdeyedata.ai/the-hidden-cost-of-manual-gauge-readings-and-the-ai-solutions/

\bibitem{b13} Beamex, ``Manual Data Entry Errors,'' 2021. [Online]. Available: https://blog.beamex.com/manual-data-entry-errors

\bibitem{b14} Ocrolus, ``8 Human Error Examples and Statistics,'' Dec. 17, 2024. [Online]. Available: https://www.ocrolus.com/blog/human-error-8-eye-popping-sets-of-stats-and-examples/

\bibitem{b15} C. D. Manning, P. Raghavan, and H. Schütze, \textit{Introduction to Information Retrieval}. Cambridge, U.K.: Cambridge Univ. Press, 2008.

\bibitem{b16} C. Mao, M. Mohri, and Y. Zhong, ``Cross-entropy loss functions: Theoretical analysis and applications,'' in \textit{Proc. Mach. Learn. Res.}, vol. 206, pp. 6069--6108, 2023.

\bibitem{b17} K. He, X. Zhang, S. Ren, and J. Sun, ``Deep residual learning for image recognition,'' in \textit{Proc. IEEE Conf. Comput. Vis. Pattern Recognit. (CVPR)}, pp. 770--778, 2016.

\bibitem{b18} L. Ntantiso, A. Bagula, O. Ajayi, and F. Kahenga-Ngongo, ``A Review of Federated Learning: Algorithms, Frameworks and Applications,'' in R. A. Saeed, A. D. Bakari, and Y. H. Sheikh, Eds., \textit{Towards new e-Infrastructure and e-Services for Developing Countries (AFRICOMM 2022)}, Lecture Notes of the Institute for Computer Sciences, Social Informatics and Telecommunications Engineering, vol. 499. Cham: Springer, 2023, pp. 1--20. doi: 10.1007/978-3-031-34896-9\_20

\bibitem{b19} M. Everingham, L. Van Gool, C. K. Williams, J. Winn, and A. Zisserman, ``The Pascal visual object classes challenge 2012 (VOC2012) results,'' 2012.

\bibitem{b20} A. R. Rizaman, H. Selamat, and N. Khamis, ``Digitization of analogue meter reading using convolution neural network,'' \textit{Mekatronika: Journal of Intelligent Manufacturing and Mechatronics}, vol. 3, no. 1, pp. 28--34, 2021.

\bibitem{b21} J. Leon-Alcazar, Y. Alnumay, C. Zheng, H. Trigui, S. Patel, and B. Ghanem, ``Learning to read analog gauges from synthetic data,'' in \textit{Proc. IEEE/CVF Winter Conf. on Applications of Computer Vision (WACV)}, pp. 8616--8625, 2024.

\end{thebibliography}
\end{document}